%% file: main.tex
\documentclass[11pt,a4paper]{article}
\usepackage[hyperref]{acl2019}
\usepackage{times}  
\usepackage{helvet}  
\usepackage{courier}  
\usepackage{url}  
\usepackage{graphicx}  

\usepackage{latexsym}
\usepackage{bm}
\usepackage{amsmath,amsfonts,amssymb,amsthm}
\usepackage{tikz}
\usetikzlibrary{arrows, shapes, decorations.pathreplacing, calc, positioning}
\usepackage{tkz-graph}
\usepackage{algorithm}
\usepackage{algpseudocode}
\usepackage{caption}
\usepackage{subcaption}
\usepackage{fancyvrb}
\usepackage{enumitem,kantlipsum}
\usepackage{tabularx}
\usepackage{multirow}

\aclfinalcopy 
\def\aclpaperid{1838} 

\title{Sentiment Tagging with Partial Labels using Modular Architectures}

\author{Xiao Zhang \\
  Purdue University \\
  {\tt zhang923@purdue.edu} \\\And
  Dan Goldwasser \\
  Purdue University \\
  {\tt dgoldwas@purdue.edu} \\}

\date{}

\begin{document}
\maketitle
\begin{abstract}


Many NLP learning tasks can be decomposed into several distinct sub-tasks, each associated with a \textit{partial} label.  In this paper we focus on a popular class of learning problems, sequence prediction applied to several sentiment analysis tasks, and  suggest a modular learning approach in which different sub-tasks are learned using separate functional modules, combined to perform the final task while sharing information. Our experiments show this approach helps constrain the learning process and can alleviate some of the supervision efforts. 


\end{abstract}

\section{Introduction}
\label{sec:intro}
\input{introduction}

\section{Related Works}
\label{sec:related}
\input{related}

\section{Architectures for Sequence Prediction }
\label{sec:sp}
\input{sp_overview}

\section{Functional Decomposition of Composite Tasks}
\label{sec:taskDecomp}
\input{task_decomposition}

\section{Learning using Full and Partial Labels}
\label{sec:training}
\input{training}

\section{Experimental Evaluation}
\label{sec:Experiments}
\input{experiments}

\section{Conclusions}
\label{sec:concl}
\input{concl}

 \section*{Acknowledgements}
 \label{sec:acknowledgements}
 We thank the reviewers for their insightful comments. We thank the NVIDIA Corporation for their GPU donation, used in this work. This work was partially funded by a Google Gift.

\bibliography{mybib}
\bibliographystyle{acl_natbib}

\appendix
\label{sec:append}
\input{appendix}



\end{document}

%% file: introduction.tex
Many natural language processing tasks attempt to replicate complex human-level judgments, which often rely on a composition of several sub-tasks into a unified judgment. 
For example, consider the Targeted-Sentiment task~\cite{Mitchell2013}, assigning a sentiment polarity score to entities depending on the context that they appear in. Given the sentence \textit{``according to a CNN poll, Green Book will win the best movie award''}, the system has to identify both entities, and associate the relevant sentiment value with each one (neutral with \textit{CNN}, and positive with \textit{Green Book}). This task can be viewed as a combination of two tasks, entity identification, locating contiguous spans of words corresponding to relevant entities, and sentiment prediction, specific to each entity based on the context it appears in.
%
%
%
Despite the fact that this form of functional task decomposition is natural for many learning tasks, it is typically ignored and learning is defined as a monolithic process, combining the tasks into a single learning problem. 

Our goal in this paper is to take a step towards modular learning architectures that exploit the learning tasks' inner structure, and as a result simplify the learning process and reduce the annotation effort. 
We introduce a novel task decomposition approach, \textit{learning with partial labels}, in which the task output labels decompose  hierarchically, into partial labels capturing different aspects, or sub-tasks, of the final task. We show that learning with partial labels can help support weakly-supervised learning  when only some of the partial labels are available.

Given the popularity of sequence labeling tasks in NLP, we demonstrate the strength of this approach over several sentiment analysis tasks, adapted for sequence prediction. These include target-sentiment prediction~\cite{Mitchell2013}, aspect-sentiment prediction~\cite{pontiki2016semeval} and  subjective text span identification and polarity prediction~\cite{hltcoe2013semeval}. To ensure the broad applicability of our approach to other problems, we extend the popular LSTM-CRF~\cite{Lample2016} model that was applied to many sequence labeling tasks\footnote{We also provide analysis for NER in the apendix}.

The modular learning process corresponds to a task decomposition, in which the prediction label, $y$, is deconstructed into a set of partial labels $\{y^0,..,y^k\}$, each defining a sub-task, capturing a different aspect of the original task.
Intuitively, the individual sub-tasks are significantly easier to learn, suggesting that if their dependencies are modeled correctly when learning the final task, they can constrain the learning problem, leading to faster convergence and a better overall learning outcome. In addition, the modular approach helps alleviate the supervision problem, as often providing full supervision for the overall task is costly, while providing additional partial labels is significantly easier.  For example, annotating entity segments syntactically is considerably easier than determining their associated sentiment, which requires understanding the nuances of the context they appear in semantically. By exploiting modularity, the entity segmentation partial labels can be used to  help improve that specific aspect of the overall task.

Our modular task decomposition approach is partially inspired by findings in cognitive neuroscience, namely the \textit{two-streams hypothesis}, a widely accepted model for neural processing of cognitive information in vision and hearing~\cite{Eysenck2005}, suggesting the brain processes information in a modular way, split between a ``where'' (dorsal) pathway, specialized for locating objects and a ``what'' (ventral) pathway, associated with object representation and recognition~\cite{Mishkin1983,Geschwind1987,Kosslyn1987,Rueckl1989}.  \citeauthor{Jacobs1991} \shortcite{Jacobs1991} provided a computational perspective, investigating the ``what'' and ``where'' decomposition on a computer vision task. We observe that this task decomposition naturally fits many NLP tasks and borrow the notation. In the target-sentiment tasks we address in this paper, the segmentation tagging task  can be considered as a ``where''-task (i.e., the location of entities), and the sentiment recognition as the ``what''-task.

Our approach is related to multi-task learning~\cite{Caruana1997}, which has been extensively applied in NLP  \cite{Toshniwal2017,Eriguchi17,Collobert2011,Luong2016,Liu2018}.
However, instead of simply aggregating the objective functions of several \textit{different} tasks, we suggest to \textit{decompose a single task into multiple inter-connected sub-tasks} and then integrate the representation learned into a single module for the final decision. We study several modular neural architectures, which differ in the way information is shared between tasks, the learning representation associated with each task and the way the dependency between decisions is modeled.

Our experiments were designed to answer two questions.  \textit{First}, can the task structure be exploited to simplify a complex learning task by using a modular approach?  \textit{Second}, can partial labels be used effectively to reduce the annotation effort? 

To answer the first question, we conduct experiments over several sequence prediction tasks, and compare our approach to several recent models for deep structured prediction~\cite{Lample2016,Ma2016,Liu2018}, and when available, previously published results~\cite{Mitchell2013,zhang2015,li2017,ma2018joint} 
We show that modular learning indeed helps simplify the learning task compared to traditional monolithic approaches. To answer the second question, we evaluate our model's ability to leverage partial labels in two ways. First, by restricting the amount of full labels, and observing the improvement when providing increasing amounts of partial labels for only one of the sub-tasks. Second, we learn the sub-tasks using completely disjoint datasets of partial labels, and show that the knowledge learned by the sub-task modules can be integrated into the final decision module using a small amount of full labels.

{\bf Our contributions:} (1) We provide a general modular framework for sequence learning tasks. While we focus on sentiment analysis task, the framework is broadly applicable to many other tagging tasks, for example, NER~\cite{Carreras2002,Lample2016} and SRL~\cite{zhou2015end}, to name a few. (2) We introduce a novel weakly supervised learning approach, \textit{learning with partial labels}, that exploits the modular structure to reduce the supervision effort. (3)  We evaluated our proposed model, in both the fully-supervised and weakly supervised scenarios, over several sentiment analysis tasks. 



%% file: related.tex
From a technical perspective, our task decomposition approach is related to multi-task learning~\cite{Caruana1997}, specifically, when the tasks share information using a shared deep representation~\cite{Collobert2011,Luong2016}. However, most prior works aggregate multiple losses on either different pre-defined tasks at the final layer~\cite{Collobert2011,Luong2016}, or on a language model at the bottom level \cite{Liu2018}. This work suggests to decompose a given task into sub-tasks whose integration comprise the original task. To the best of our knowledge, \citeauthor{ma2018joint} \shortcite{ma2018joint}, focusing on targeted sentiment is most similar to our approach. They suggest a joint learning approach, modeling a sequential relationship between two tasks, entity identification and target sentiment. We take a different approach viewing each of the model components as a separate module, predicted independently and then integrated into the final decision module. As we demonstrate in our experiments, this approach leads to better performance and increased flexibility, as it allows us to decouple the learning process and learn the tasks independently.  
%
%
Other modular neural architectures were recently studied for tasks combining vision and language analysis~\cite{Andreas2016,Hu2017,Yu2018}, and were tailored for the grounded language setting.  To help ensure the broad applicability of our framework, we provide a general modular network formulation for sequence labeling tasks by adapting a neural-CRF to capture the task structure.  This family of models, combining structured prediction with deep learning showed promising results~\cite{Gillick2015,Lample2016,Ma2016,zhang2015,li2017}, by using rich representations through neural models to generate decision candidates, while utilizing an inference procedure to ensure coherent decisions. Our main observation is that modular learning can help alleviate some of the difficulty involved in training these powerful models. 

%% file: sp_overview.tex
Using neural networks to generate emission potentials in CRFs was applied successfully in several sequence  prediction tasks, such as word segmentation \cite{Chen2017}, NER \cite{Ma2016,Lample2016}, chunking and PoS tagging \cite{Liu2018,Zhang2017}. A sequence is represented as a sequence of $L$ tokens: $\bm{x} = [x_1, x_2,\dots, x_L]$, each token corresponds to a label $y\in \mathcal{Y}$, where $\mathcal{Y}$ is the set of all possible tags. An inference procedure is designed to find the most probable sequence $\bm{y}^{*} = [y_1, y_2,\dots, y_L]$ by solving, either exactly or approximately, the following optimization problem:
$$
\bm{y}^{*} = \arg\max_{\bm{y}}P(\bm{y}|\bm{x}).
$$
Despite the difference in tasks, these models follow a similar general architecture: 
(1) Character-level information, such as prefix, suffix and capitalization, is represented through a character embedding layer learned using a bi-directional LSTM (BiLSTM). (2) Word-level information is obtained through a word embedding layer. (3) The two representations are concatenated to represent an input token, used as input to a word-level BiLSTM which generates the emission potentials for a succeeding CRF. (4) The CRF is used as an inference layer to generate the globally-normalized probability of possible tag sequences.

\subsection{CRF Layer}
A CRF model describes the probability of predicted labels $\bm{y}$, given a sequence $\bm{x}$ as input, as
\begin{equation}
P_{\bm{\Lambda}}(\bm{y}|\bm{x}) = \frac{e^{\Phi(\bm{x}, \bm{y})}}{Z}, \notag
\end{equation}
where $Z = \sum\limits_{\tilde{\bm{y}}}e^{\Phi(\bm{x}, \tilde{\bm{y}})}$ is the partition function that marginalize over all possible assignments to the predicted labels of the sequence, and $\Phi(\bm{x}, \bm{y})$ is the scoring function, which is defined as: 
\begin{equation}
\Phi(\bm{x}, \bm{y}) = \sum\limits_{t}\phi(\bm{x}, y_t) + \psi(y_{t-1}, y_t). \notag
\end{equation}
The partition function $ Z $ can be computed efficiently via the forward-backward algorithm. The term $\phi(\bm{x}, y_t)$ corresponds to the score of a particular tag $y_t$ at position $t$ in the sequence, and $\psi(y_{t-1}, y_t)$ represents the score of transition from the tag at position $t-1$ to the tag at position $t$. In the Neural CRF model, $\phi(\bm{x}, y_t)$ is generated by the aforementioned Bi-LSTM while $\psi(y_{t-1}, y_t)$ by a transition matrix.

%% file: task_decomposition.tex
To accommodate our task decomposition approach, we first define the notion of partial labels, and then discuss different neural architectures capturing the dependencies between the modules trained over the different partial labels.

\noindent\textbf{Partial Labels and Task Decomposition:}
Given a learning task, defined over an output space $y\in \mathcal{Y}$, where $\mathcal{Y}$ is the set of all possible tags, each specific label $y$ is decomposed into a set of partial labels, $\{y^0,..,y^k\}$. We refer to $y$ as the \textit{full} label. According to this definition, a specific assignment to all $k$ partial labels defines a single full label.
 ~Note the difference between \textit{partially labeled data}~\cite{cour2011learning}, in which instances can have more than a single full label, and our setup in which the labels are partial.

\begin{figure}[ht]
	\centering
	\includegraphics[width=0.485\textwidth]{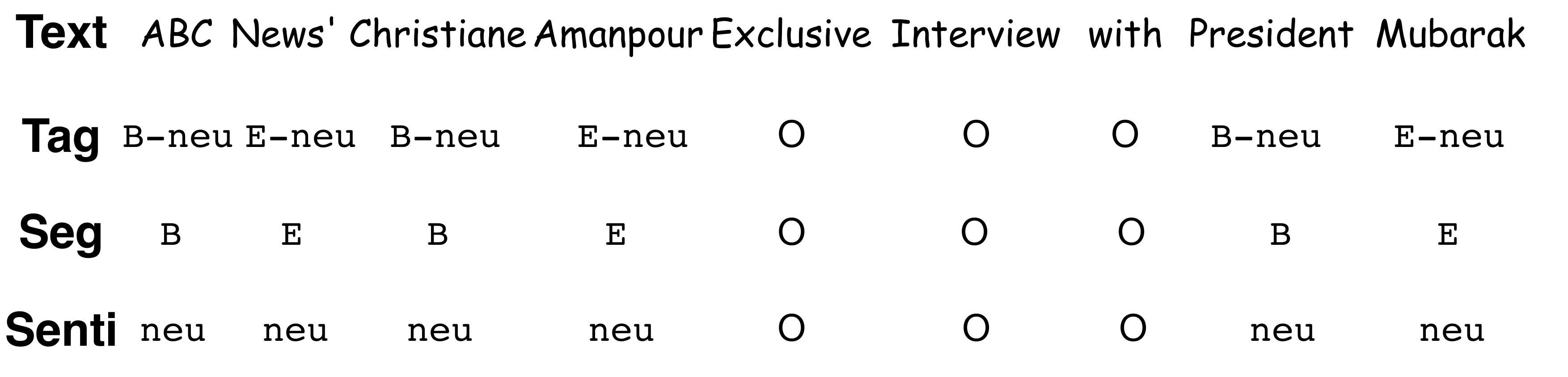}
	\caption{Target-sentiment decomposition example.}
	\label{tarsentieg}
\end{figure}

In all our experiments, the partial labels refer to two sub-tasks, (1) a segmentation task, identifying \textit{Beginning}, \textit{Inside} and \textit{Outside} of an entity or aspect. (2) one or more type recognition tasks, recognizing the aspect type and/or the sentiment polarity associated with it. Hence, a tag $y_t$ at location $t$ is divided into $y_{t}^{seg}$ and $y_{t}^{typ}$, corresponding to segmentation and type (sentiment type here) respectively.  Fig. \ref{tarsentieg} provides an example of the target-sentiment task. Note that the sentiment labels do not capture segmentation information.


\noindent\textbf{Modular Learning architectures: }
We propose three different models, in which information from the partial labels can be used. All the models have similar modules types, corresponding to the \textit{segmentation} and \textit{type} sub-tasks, and the decision module for predicting the final task. The modules are trained over the partial segmentation  ($\bm{y}^{seg}$) and type ( $\bm{y}^{typ}$) labels, and the full label $\bm{y}$ information, respectively. 

\begin{figure*}[ht]
\centering
\begin{subfigure}{0.26\textwidth}
\includegraphics[width=1\textwidth]{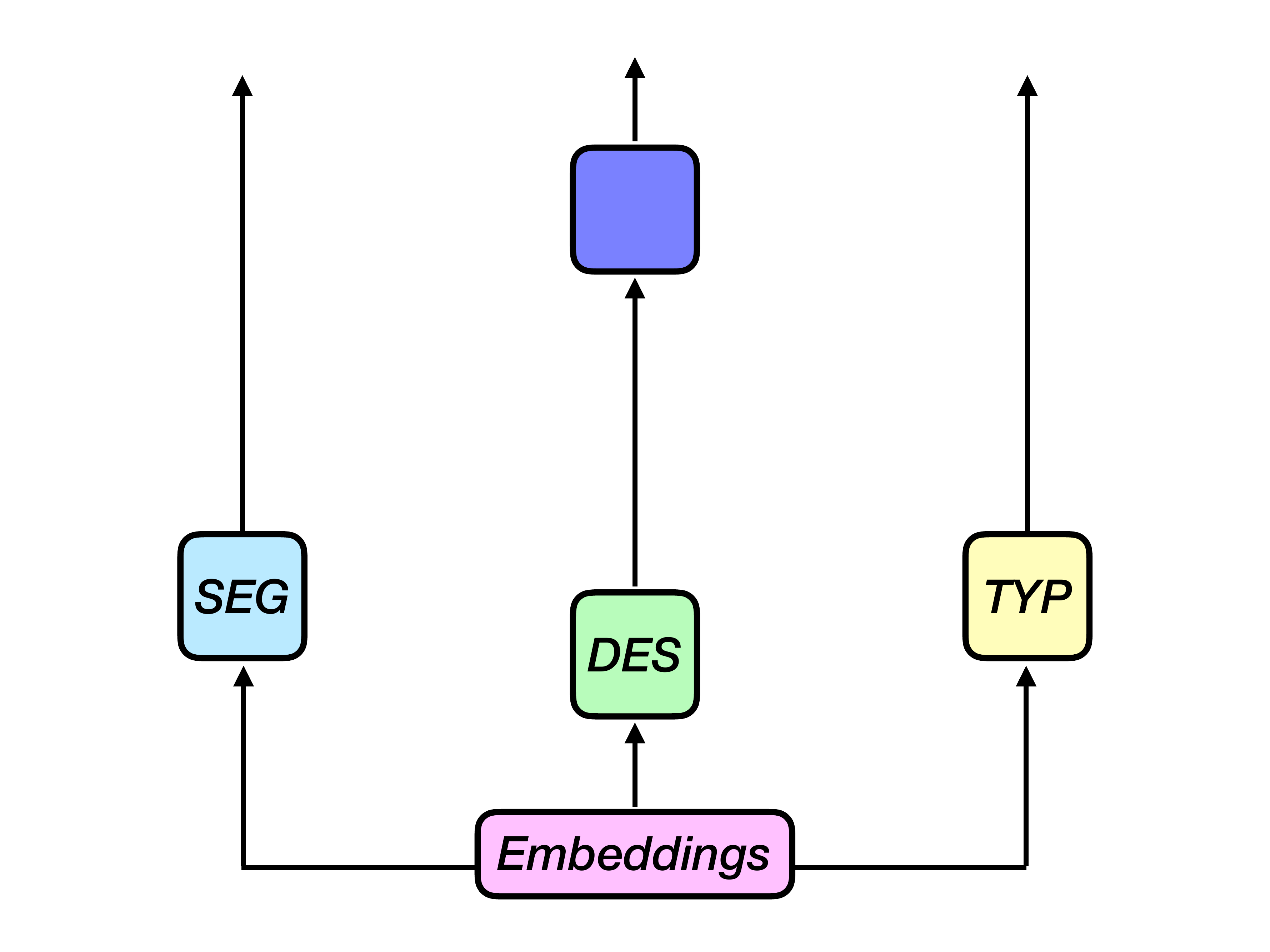}
\caption{LSTM-CRF-T}
\label{model1}
\end{subfigure}~~~
\hspace{1em}
\begin{subfigure}{0.26\textwidth}
\includegraphics[width=1\textwidth]{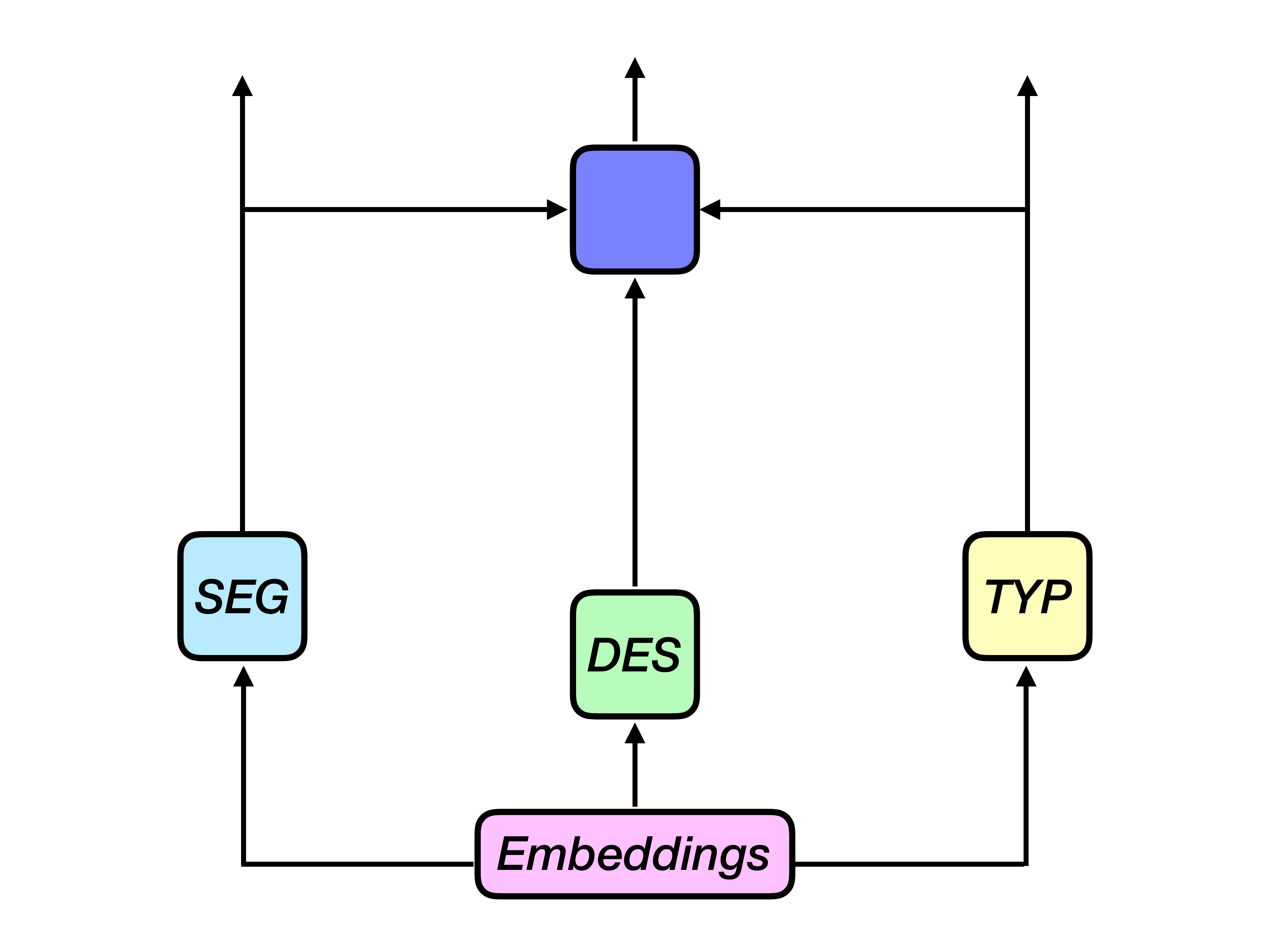}
\caption{LSTM-CRF-TI}
\label{model2}
\end{subfigure}~~~
\hspace{1em}
\begin{subfigure}{0.26\textwidth}
\includegraphics[width=1\textwidth]{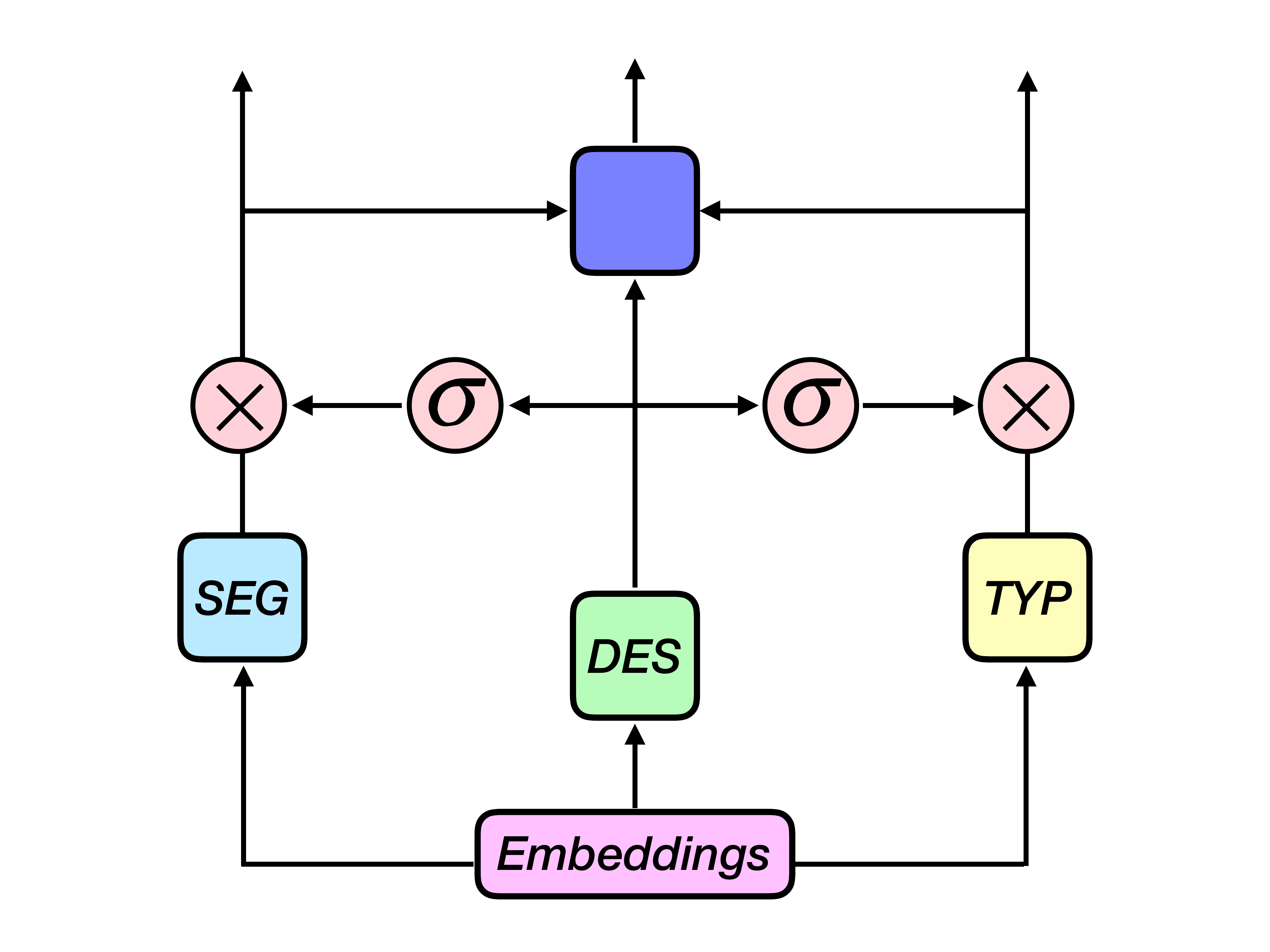}
\caption{LSTM-CRF-TI (G)}
\label{model3}
\end{subfigure}
\caption{Three modular models for task decomposition. In them, blue blocks are \textit{segmentation} modules, detecting entity location and segmentation, and yellow blocks are the \textit{type} modules, recognizing the entity type or sentiment polarity. Green blocks are the final decision modules, integrating all the decisions. (G) refers to ``Guided Gating"}
\label{models}
\end{figure*}

These three models differ in the way they share information. \textbf{Model 1}, denoted \textit{Twofold Modular, LSTM-CRF-T}, is similar in spirit to multi-task learning~\cite{Collobert2011} with three separate modules. \textbf{Model 2}, denoted \textit{Twofold modular Infusion, (LSTM-CRF-TI}) and \textbf{Model 3}, denoted \textit{Twofold modular Infusion with guided gating, (LSTM-CRF-TI(g)}) both infuse information flow from two sub-task modules into the decision module. The difference is whether the infusion is direct or goes through a guided gating mechanism. The three models are depicted in Fig.~\ref{models} and described in details in the following paragraphs. In all of these models, underlying neural architecture are used for the emission potentials when CRF inference layers are applied on top.
%
 
\subsection{Twofold Modular Model}

The twofold modular model enhances the original monolithic model by using multi-task learning with shared underlying representations. The segmentation module and the type module are trained jointly with the decision module, and all the modules share information by using the same embedding level representation, as shown in Figure \ref{model1}. Since the information above the embedding level is independent, the LSTM layers in the different modules do not share information, so we refer to these layers of each module as  \textit{private}.
 
The segmentation module predicts the segmentation BIO labels at position $t$ of the sequence by using the representations extracted from its private word level bi-directional LSTM (denoted as $\mathcal{H}^{seg}$) as emission for a individual CRF:
\begin{align*}
\bm{h}_{t}^{seg} = \mathcal{H}^{seg}(\bm{e}_{t}, \overrightarrow{\bm{h}}_{t-1}^{seg}, \overrightarrow{\bm{h}}_{t+1}^{seg}),\\
\phi(\bm{x}, y^{seg}_{t}) = \bm{W}^{seg\intercal}\bm{h}^{seg}_{t} + \bm{b}^{seg},
\end{align*}
where $\bm{W}^{seg}$ and $\bm{b}^{seg}$  denote the parameters of the segmentation module emission layer, and $\mathcal{H}^{seg}$ denotes its private LSTM layer.


This formulation allows the model to forge the segmentation path privately through back-propagation by providing the segmentation information $\bm{y}^{seg}$ individually, in addition to the complete tag information $\bm{y}$. 

The type module, using  $\bm{y}^{typ}$, is constructed in a similar way.
By using representations from the its own private LSTM layers, the type module predicts the sentiment (entity) type at position $t$ of the sequence :
\begin{align*}
\bm{h}^{typ}_{t} = \mathcal{H}^{typ}(\bm{e}_{t}, \overrightarrow{\bm{h}}_{t-1}^{typ}, \overrightarrow{\bm{h}}_{t+1}^{typ}),\\
\phi(\bm{x}, y^{typ}_{t}) = \bm{W}^{typ\intercal}\bm{h}^{typ}_{t} + \bm{b}^{typ}.
\end{align*} 


Both the segmentation information $\bm{y}^{seg}$ and the type information $\bm{y}^{typ}$ are provided together with the complete tag sequence $\bm{y}$, enabling the model to learn segmentation and type recognition simultaneously using two different paths. Also, the decomposed tags naturally augment \textit{more training data} to the model, avoiding over-fitting due to more complicated structure. The shared representation beneath the private LSTMs layers are updated via the back-propagated errors from all the three modules.


\subsection{Two-fold Modular Infusion Model}
The twofold modular infusion model provides a stronger connection between the functionalities of the two sub-tasks modules and the final decision module, differing from multi-task leaning. 
%

In this model, instead of separating the pathways from the decision module as in the previous twofold modular model, the segmentation and the type representation are used as input to the final decision module. The model structure is shown in Figure \ref{model2}, and can be described formally as:  
\begin{align*}
\bm{I}_{t}^{seg} = \bm{W}^{seg\intercal}\bm{h}_{t}^{seg} + \bm{b}^{seg},\\
\bm{I}_{t}^{typ} = \bm{W}^{typ\intercal}\bm{h}_{t}^{typ} + \bm{b}^{typ},\\
S_{t} = \bm{W}^{\intercal}[\bm{h}_{t};\bm{I}_{t}^{seg};\bm{I}_{t}^{typ}] + \bm{b},
\end{align*}
where $S_{t}$ is the shared final emission potential to the CRF layer in the decision module, and $;$ is the concatenation operator, combining the representation from the decision module and that from the type module and the segmentation module.

The term \textit{``Infusion''} used for naming this module is intended to indicate that both modules actively participate in the final decision process, rather than merely form two independent paths as in the twofold modular model. This formulation provides an alternative way of integrating the auxiliary sub-tasks back into the major task in the neural structure to help improve learning.

\subsection{Guided Gating Infusion}
In the previous section we described a way of infusing information from other modules naively by simply concatenating them. But intuitively, the hidden representation from the decision module plays an important role as it is directly related to the final  task we are interested in. To effectively use the information from other modules forming sub-tasks, we design a gating mechanism to dynamically control the amount of information flowing from other modules by infusing the expedient part while excluding the irrelevant part, as shown in Figure \ref{model3}. This gating mechanism uses the information from the decision module to guide the information from other modules, thus we name it as guided gating infusion, which we describe formally as follows:
\begin{align*}
\bm{I}_{t}^{seg} = &\sigma(\bm{W_1}h_{t}+\bm{b_1})\otimes(\bm{W}^{seg\intercal}\bm{h}_{t}^{seg} + \bm{b}^{seg}),\\
\bm{I}_{t}^{typ} = &\sigma(\bm{W_2}h_{t}+\bm{b_2})\otimes(\bm{W}^{typ\intercal}\bm{h}_{t}^{typ} + \bm{b}^{typ}),\\
S_{t} = &\bm{W}^{\intercal}[\bm{h}_{t};\bm{I}_{t}^{seg};\bm{I}_{t}^{typ}] + \bm{b},
\end{align*}
where $\sigma$ is the logistic sigmoid function and $\otimes$ is the element-wise multiplication. The $\{W_1, W_2, b_1, b_2\}$ are the parameters of these guided gating, which are updated during the training to maximize the overall sequence labeling performance.

%% file: training.tex
Our objective naturally rises from the model we described in the text. Furthermore, as our experiments show, it is easy to generalize this objective, to a ``semi-supervised" setting, in which the learner has access to only a few fully labeled examples and additional partially labeled examples. E.g., if only segmentation is annotated but the type information is missing.
The loss function is a linear combination of the negative log probability of each sub-tasks, together with the decision module:
\par\nobreak{\small
\begin{align}
\mathcal{J} = & -\sum\limits_{i}^{N} \log P(\bm{y}^i|\bm{x}^i) + \alpha \log P(\bm{y}^{seg(i)}|\bm{x}^{(i)})\notag\\
&+ \beta \log P(\bm{y}^{typ(i)}|\bm{x}^{(i)}),
\label{tloss}
\end{align}
}
where $N$ is the number of examples in the training set, $\bm{y}^{seg}$ and $\bm{y}^{typ}$ are the decomposed segmentation and type tags corresponding to the two sub-task modules, and $\alpha$ and $\beta$ are the hyper-parameters controlling the importance of the two modules contributions respectively. 

If the training example is fully labeled with both segmentation and type annotated,  training is straightforward; if the training example is partially labeled, e.g., only with segmentation but without type, we can set the log probability of the type module and the decision module $0$ and only train the segmentation module. This formulation provides extra flexibility of using partially annotated corpus together with fully annotated corpus to improve the overall performance.

%% file: experiments.tex

Our experimental evaluation is designed to evaluate the two key aspects of our model:

\noindent(Q1) \textit{Can the modular architecture alleviate the difficulty of learning the final task?}  To answer this question, we compare our modular architecture to the traditional neural-CRF model and several recent competitive models for sequence labeling combining inference and deep learning. The results are summarized in Tables~\ref{target-senti-main}-\ref{subject-senti-main}. 

\noindent(Q2) \textit{Can partial labels be used effectively as a new form of weak-supervision?} To answer this question we compared the performance of the model when trained using disjoint sets of partial and full labels, and show that adding examples only associated with partial labels, can help boost performance on the final task. The results are summarized in Figures~\ref{modular_ts}-\ref{domaintransfer}.

\subsection{Experimental Settings}
\subsubsection{Datasets} 
We evaluated our models over three different sentiment analysis tasks adapted for sequence prediction. We included additional results for multilingual NER in the Appendix for reference.

\paragraph{Target Sentiment Datasets}
We evaluated our models on the targeted sentiment dataset released by \citeauthor{Mitchell2013} \shortcite{Mitchell2013}, which consists of entity and sentiment annotations on both English and Spanish tweets. Similar to  previous studies \cite{Mitchell2013,zhang2015,li2017}, our task focuses on people and organizations (collapsed into \textit{volitional named entities} tags) and the sentiment associated with their description in tweets. After this processing, the labels of each tweets are  composed of both segmentation (entity spans) and types (sentiment tags). 

We used the original 10-fold cross validation splits to calculate averaged F1 score, using 10\% of the training set for development. We used the same metrics in \citeauthor{zhang2015}~\shortcite{zhang2015} and \citeauthor{li2017}~\shortcite{li2017} for a fair comparison.

\paragraph{Aspect Based Sentiment Analysis Datasets}  We used the Restaurants dataset provided by
SemEval 2016 Task 5 subtask 1, consisting of opinion target (aspect) expression segmentation,  aspect classification and matching sentiment prediction. In the original task definition, the three tasks were designed as a pipeline, and assumed gold aspect labels when predicting the matching sentiment labels. Instead, our model deals with the challenging end-to-end setting by casting the problem as a sequence labeling task, labeling each aspect segment with the aspect label and sentiment polarity\footnote{using only the subset of the data containing sequence information}.


\paragraph{Subjective Polarity Disambiguation Datasets}
We adapted the SemEval 2013 Task 2 subtask A as another task to evaluate our model. In this task, the system is given a marked phrase inside a longer text, and is asked to label its polarity. Unlike the original task, we did not assume the sequence is known, resulting in two decisions, identifying subjective expressions (i.e., a segmentation task) and labeling their polarity, which can be modeled jointly as a sequence labeling task.



\subsubsection{Input Representation and Model Architecture}
Following previous studies \cite{Ma2016,Liu2018} showing that the word embedding choice can significantly influence performance, 
%
we used the pre-trained GloVe 100 dimension Twitter embeddings only for all tasks in the main text.
All the words not contained in these embeddings (OOV, out-of-vocabulary words) are treated as an ``unknown'' word. 
Our models were deployed with minimal hyper parameters tuning, and can be briefly summarized as: the character embeddings has dimension 30, the hidden layer dimension of the character level LSTM is 25, and the hidden layer of the word level LSTM has dimension 300. Similar to \citeauthor{Liu2018} \shortcite{Liu2018}, we also applied highway networks~\cite{Srivastava2015} from the character level LSTM to the word level LSTM. In our pilot study, we shrank the number of parameters in our modular architectures to around one third such that the total number of parameter is similar as that in the LSTM-CRF model, but we did not observe a significant performance change so we kept them as denoted.
The values of $\alpha$ and $\beta$ in the objective function were always set to 1.0.

\subsubsection{Learning}
We used \textsc{BIOES} tagging scheme but only during the training and convert them back to \textsc{BIO2} for evaluation for all tasks\footnote{Using BIOES improves model complexity in Training, as suggested in previous studies. But to make a fair comparison to most previous work, who used BIO2 for evaluation, we converted labels to BIO2 system in the testing stage. (To be clear, using BIOES in the testing actually yields higher f1 scores in the testing stage, which some previous studies used unfairly)}. Our model was implemented using \textit{pytorch}~\cite{Paszke2017}. To help improve performance we parallelized the forward algorithm and the Viterbi algorithm on the GPU. All the experiments were run on NVIDIA GPUs. We used the Stochastic Gradient Descent (SGD) optimization of batch size 10, with a momentum 0.9 to update the model parameters, with the learning rate 0.01, the decay rate 0.05; The learning rate decays over epochs by $\eta/(1+e*\rho)$, where $\eta$ is the learning rate, $e$ is the epoch number, and $\rho$ is the decay rate. We used gradient clip to force the absolute value of the gradient to be less than 5.0. We used early-stop to prevent over-fitting, with a patience of 30 and at least 120 epochs. In addition to dropout, we used \textit{Adversarial Training} (AT) \cite{Goodfellow2014}, to regularize our model as the parameter numbers increase with modules. AT improves robustness to small worst-case perturbations by computing the gradients of a loss function w.r.t. the input. 
In this study, $\alpha$ and $\beta$ in Eq. \ref{tloss} are both set to $1.0$, and we leave other tuning choices for future investigation. 

\subsection{Q1: \textit{Monolithic vs. Modular Learning }}
Our first set of results are designed to compare our modular learning models, utilize partial labels decomposition, with traditional monolithic models, that learn directly over the full labels. In all three tasks, we compare with strong sequence prediction models, including LSTM-CRF~\cite{Lample2016}, which is directly equivalent to our baseline model (i.e., final task decision without the modules), and LSTM-CNN-CRF~\cite{Ma2016} and LSTM-CRF-LM~\cite{Liu2018} which use a richer latent representation for scoring the emission potentials.

\paragraph{Target Sentiment task} The results  are summarized in Tab.~\ref{target-senti-main}. We also compared our models with recently published state-of-the-art models on these datasets. To help ensure a fair comparison with ~\citeauthor{ma2018joint} which does not use inference, we also included the results of our model without the CRF layer (denoted LSTM-Ti(g)).  All of our models beat the state-of-the-art results by a large margin. The source code and experimental setup are available online\footnote{\url{https://github.com/cosmozhang/Modular_Neural_CRF}}.

\begin{table}[!ht]
\small
\centering
\begin{tabular}{ |c|c|c|c| }
\hline
System & Architecture & Eng. & Spa. \\ 
\hline
\multirow{4}{*}{\citeauthor{zhang2015} \shortcite{zhang2015}} 
 & Pipeline  & 40.06 & 43.04\\
 & Joint  & 39.67 & 43.02\\
 & Collapsed  & 38.36 & 40.00\\
 \hline
\multirow{4}{*}{\citeauthor{li2017} \shortcite{li2017}} 
 & SS & 40.11 &  42.75\\
 & +embeddings  & 43.55  & 44.13\\
 & +POS tags  & 42.21  & 42.89\\
 & +semiMarkov  & 40.94 & 42.14\\ 
 \hline
  {\citeauthor{ma2018joint} \shortcite{ma2018joint}}  & HMBi-GRU & 42.87 & 45.61 \\
 \hline
 \hline
 baseline & LSTM-CRF & 49.89 & 48.84 \\
 \hline
\multirow{4}{*}{\textit{This work}} 
 & LSTM-Ti(g)  & 45.84  & 46.59\\
 & LSTM-CRF-T  & 51.34  & 49.47\\
 & LSTM-CRF-Ti  & 51.64  & 49.74\\
 & LSTM-CRF-Ti(g)  & \textbf{52.15} &  \textbf{50.50}\\
\hline
\end{tabular}
\caption{Comparing our models with the competing models on the target sentiment task. The results are on the full prediction of both segmentation and sentiment.}
\label{target-senti-main}
\end{table}

\paragraph{Aspect Based Sentiment}
We evaluated our models on two tasks: The first uses two modules, for identifying the position of the aspect in the text (i.e., chunking) and the aspect category prediction (denoted E+A). The second adds a third module that predicts the sentiment polarity associated with the aspect (denoted E+A+S). I.e., for a given sentence, label its entity span, the aspect category of the entity and the sentiment polarity of the entity at the same time. The results over four languages are summarized in Tab.~\ref{compa_Aspect_Senti}. In all cases, our modular approach outperforms all monolithic approaches. 

\begin{table*}[!ht]
\small
\centering
\begin{tabular}{|l|c|c|c|c|c|c|c|c|}
\hline
\multirow{2}{*}{Models} & \multicolumn{2}{c|}{English} & \multicolumn{2}{c|}{Spanish} & \multicolumn{2}{c|}{Dutch} & \multicolumn{2}{c|}{Russian} \\
\cline{2-9}
 & E+A & E+A+S & E+A & E+A+S & E+A & E+A+S & E+A & E+A+S \\
\hline
LSTM-CNN-CRF\cite{Ma2016} & 58.73 & 44.20 & 64.32 & 50.34 & 51.62 & 36.88 & 58.88 & 38.13 \\
\hline 
LSTM-CRF-LM\cite{Liu2018} & 62.27 & 45.04 & 63.63 & 50.15 & 51.78 & 34.77 & 62.18 & 38.80\\
\hline
\hline
LSTM-CRF & 59.11 & 48.67 & 62.98 & 52.10 & 51.35 & 37.30 & 63.41 & 42.47\\
\hline
LSTM-CRF-T & 60.87 & 49.59 & 64.24 & 52.33 & 52.79 & 37.61 & 64.72 & 43.01\\
\hline
LSTM-CRF-TI & 63.11 & 50.19 & 64.40 & 52.85 & 53.05 & 38.07 & 64.98  & 44.03\\
\hline
LSTM-CRF-TI(g) & \textbf{64.74} & \textbf{51.24} & \textbf{66.13} & \textbf{53.47} & \textbf{53.63} & \textbf{38.65} & \textbf{65.64} & \textbf{45.65} \\
\hline
\end{tabular}
\caption{Comparing our models with recent results on the Aspect Sentiment datasets.}
\label{compa_Aspect_Senti}
\end{table*}

\paragraph{Subjective Phrase Identification and Classification }
This dataset contains tweets annotated with sentiment phrases, used for training the models. As in the original SemEval task, it is tested in two settings, in-domain, where the test data also consists of tweets, and out-of-domain, where the test set consists of SMS text messages.
%
We present the results of experiments on these data set in Table \ref{subject-senti-main}.

\begin{table}[!ht]
\small
\centering
\begin{tabular}{ |c|c|c| }
\hline
Models & Tweets & SMS \\ 
\hline
LSTM-CNN-CRF  & 35.82 & 23.23 \\
LSTM-CRF-LM  & 35.67 & 23.25 \\
\hline
\hline
LSTM-CRF & 34.15 & 26.28 \\
\hline
LSTM-CRF-T  & 35.37  & 27.11\\
LSTM-CRF-Ti  & 36.52  & 28.05\\
LSTM-CRF-Ti(g)  & \textbf{37.71} &  \textbf{29.24}\\
\hline
\end{tabular}
\caption{Comparing our models with competing models on the subjective sentiment task.}
\label{subject-senti-main}
\end{table}
\subsection{Q2: \textit{Partial Labels as Weak Supervision}}
Our modular architecture is a natural fit for learning with \textit{partial labels}. Since the modular architecture decomposes the final task into sub-tasks, the absence of certain partial labels is permitted. In this case, only the module corresponding to the available partial labels will be  updated while the other parts of the model stay fixed. 

This property can be exploited to reduce the supervision effort by defining semi-supervised learning protocols that use partial-labels when the full labels are not available, or too costly to annotate. E.g., in the target sentiment task, segmentation labels are significantly easier to annotate. 

To demonstrate this property we conducted  two sets of experiments. The first investigates how the decision module can effectively \textit{integrate} the knowledge independently learned by sub-tasks modules using different partial labels. We quantify this ability by providing varying amounts of full labels to support the integration process.  The second set studies the traditional semi-supervised settings, where we have a handful of full labels, but we have  a larger amount of partial labels.

\paragraph{Modular Knowledge Integration}
The modular architecture allows us to train each model using data obtained separately for each task, and only use a handful of examples annotated for the final task in order to integrate the knowledge learned by each module into a unified decision.  We simulated these settings by dividing the training data into three folds. We associated each one of the first two folds with the two sub-task modules. Each one of the these folds only included the partial labels relevant for that sub-task. We then used gradually increasing amounts of the third fold, consisting of the full labels, for training the decision module.

\begin{figure}[!tb]
\begin{subfigure}{0.22\textwidth}
\centering
\includegraphics[scale=0.115]{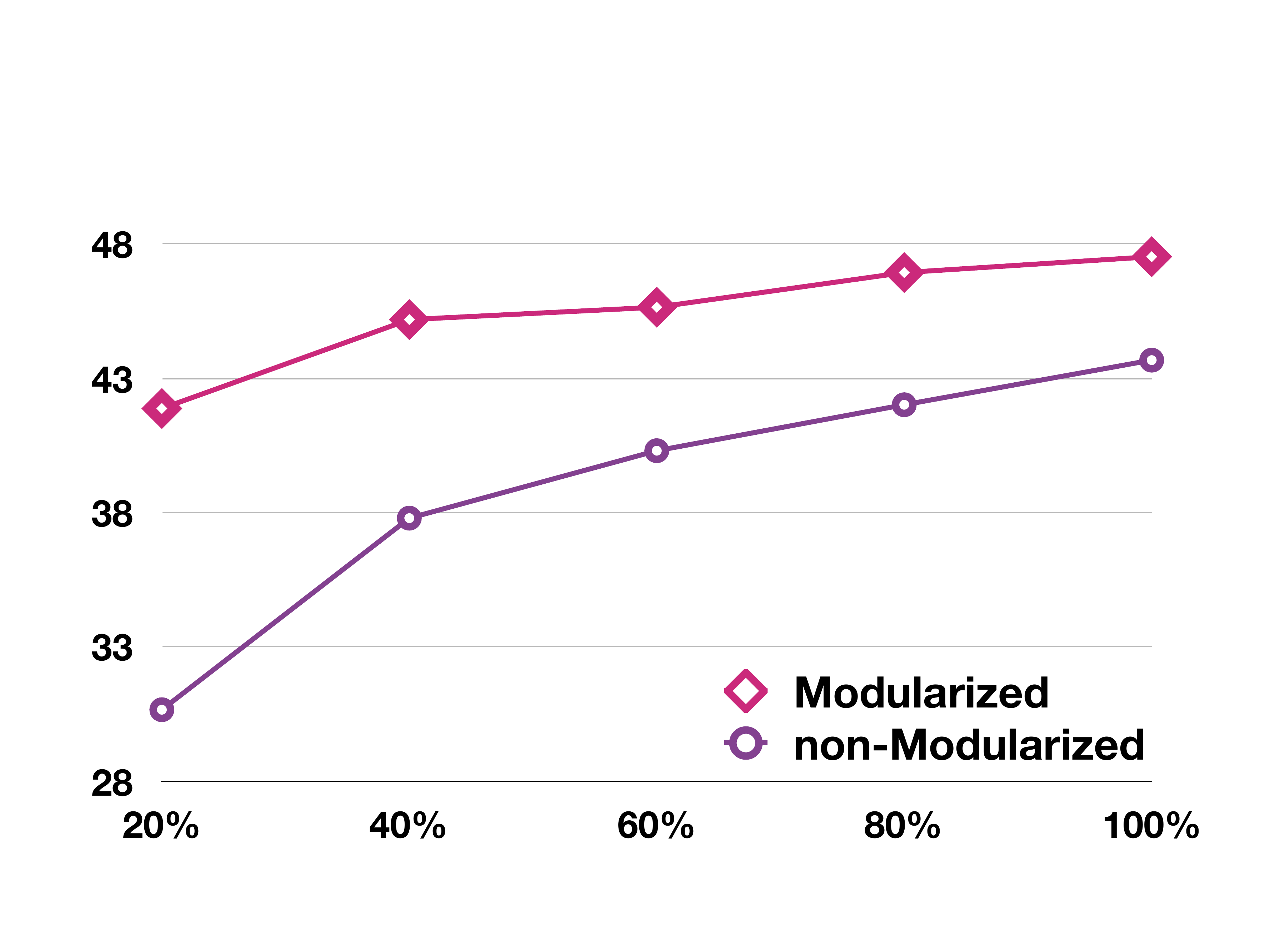}
\caption{Spanish}
\end{subfigure}
\quad
\begin{subfigure}{0.22\textwidth}
\centering
\includegraphics[scale=0.115]{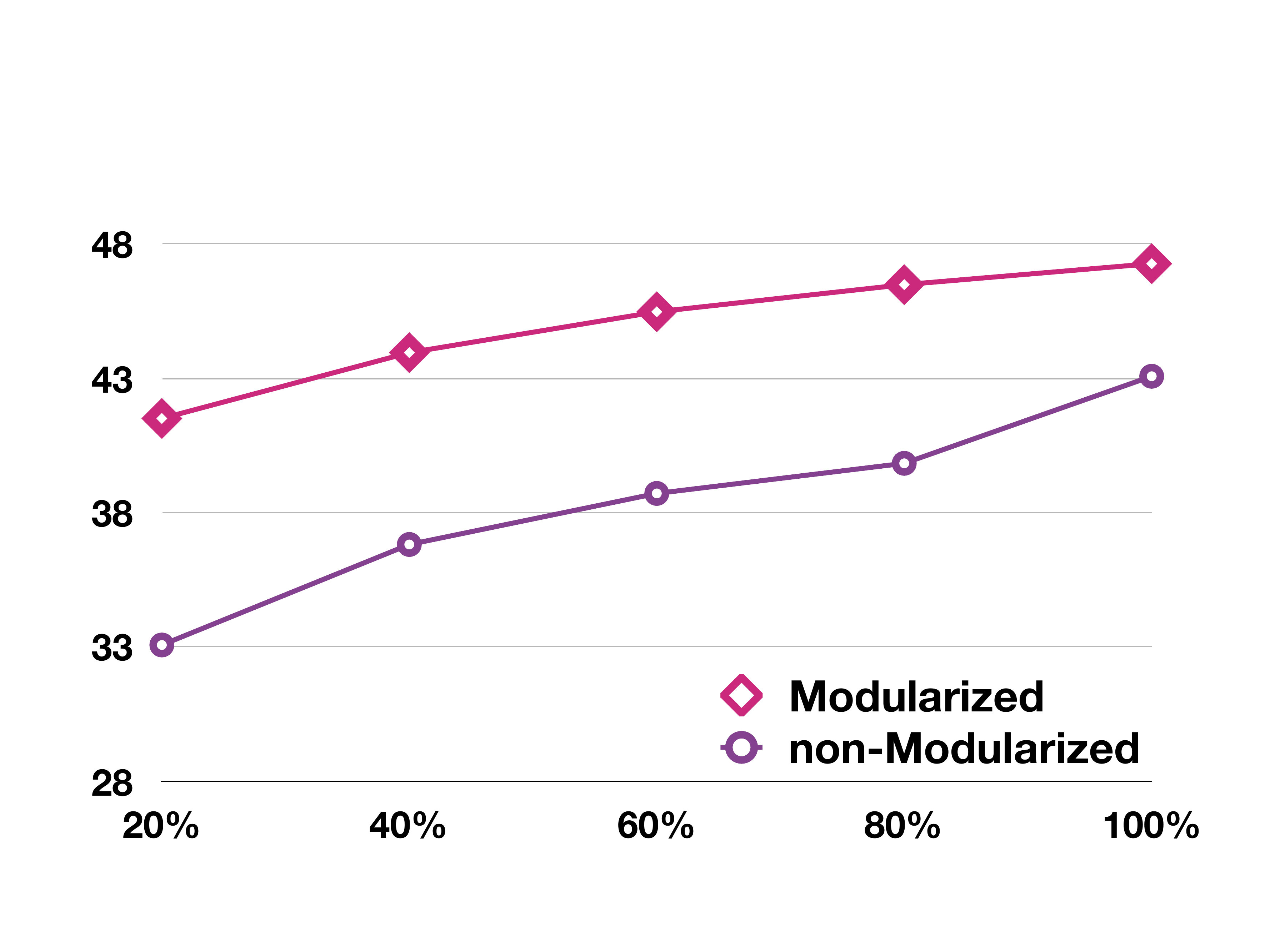}
\caption{English}
\end{subfigure}
\caption{Modular knowledge integration results on the Target Sentiment Datasets. The x-axis is the amount of percentage of the third fold of full labels. The ``non-modularized" means we only provide fully labeled data from the third fold.}
\label{modular_ts}
\end{figure}

Fig.~\ref{modular_ts} describes the outcome for target-sentiment, comparing a non-modular model using only the full labels, with the modular approach, which uses the full labels for knowledge integration. Results show that even when very little full data is available results significantly improve. Additional results show the same pattern for subjective phrase identification and classification are included in the Appendix.



\begin{figure}[!tb]
\begin{subfigure}{0.22\textwidth}
\centering
\includegraphics[scale=0.115]{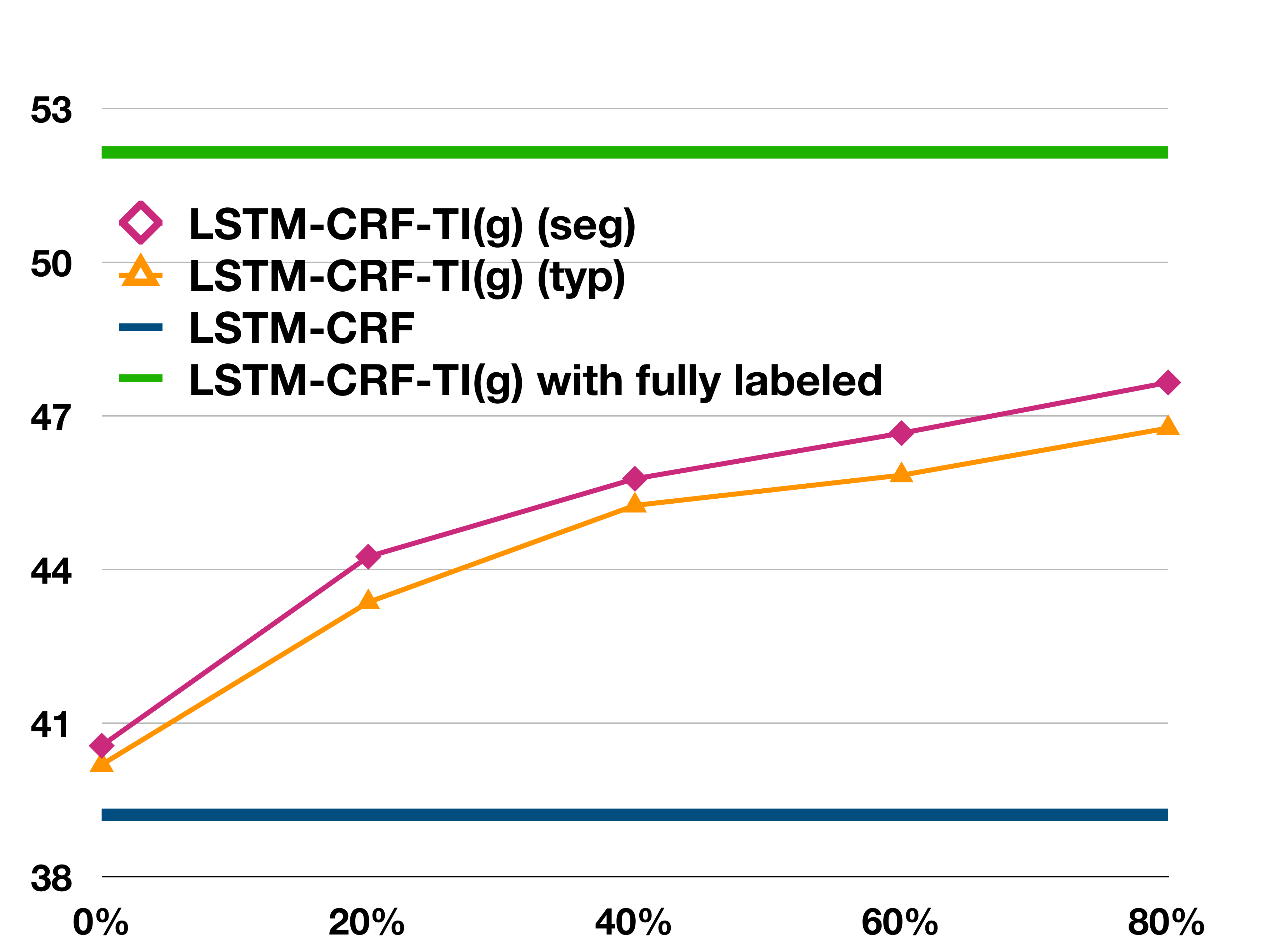}
\caption{Spanish}
\end{subfigure}
\quad
\begin{subfigure}{0.22\textwidth}
\centering
\includegraphics[scale=0.115]{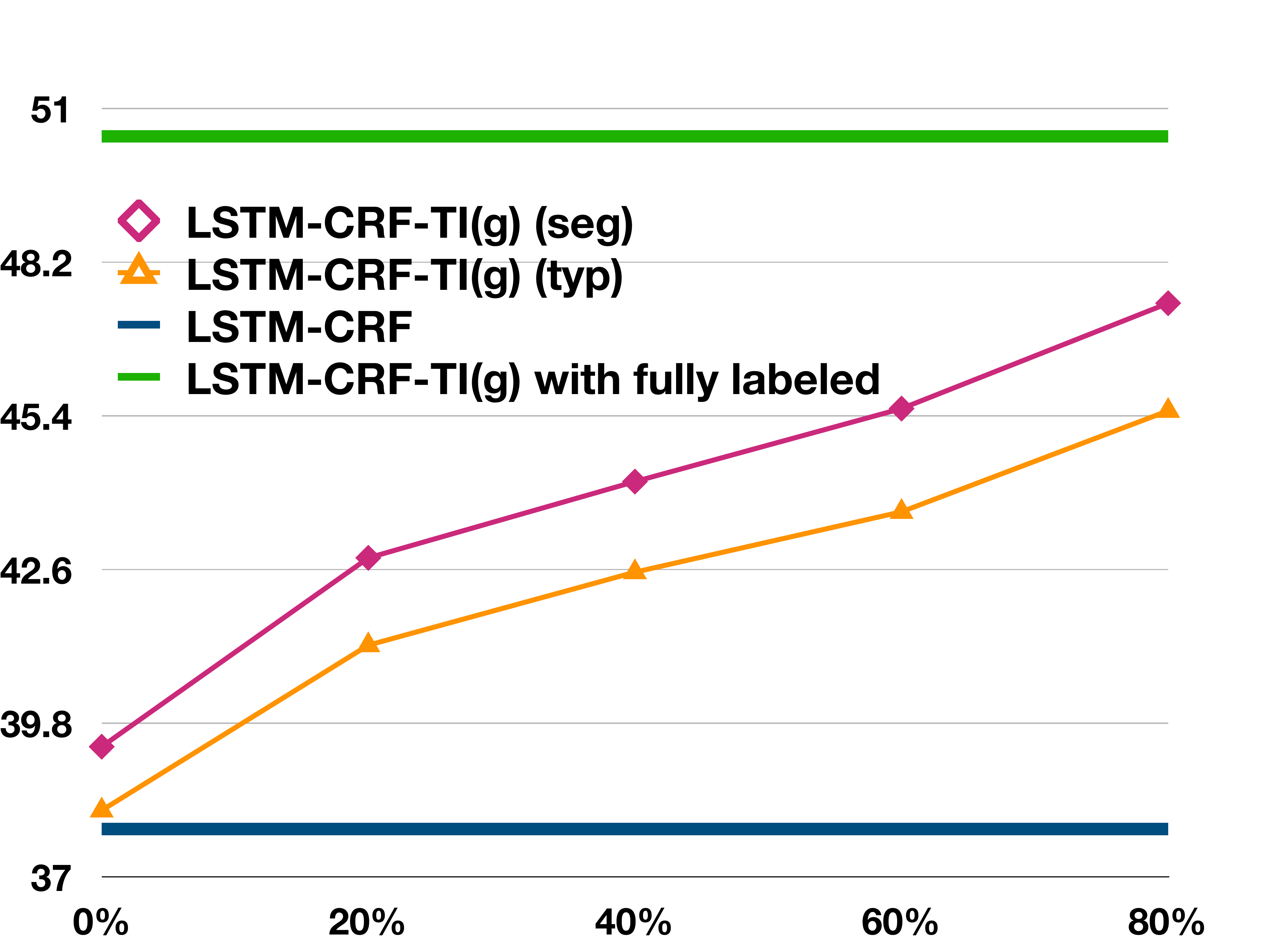}
\caption{English}
\end{subfigure}

\caption{The fully labeled data was fixed to $20\%$ of the whole training set, and gradually adding data with only segmentation information (Magenta), or with only type information (Orange), and test our model on the full prediction test. The LSTM-CRF model can only use fully labeled data as it does not decompose the task.}
\label{semi_senti}
\end{figure}

\paragraph{Learning with Partially Labeled Data}
Partially-labeled data can be cheaper and easier to obtain, especially for low-resource languages. In this set of experiments, we model these settings over the target-sentiment task. The results are summarized in Fig.~\ref{semi_senti}. We fixed the amount of full labels to $20\%$ of the training set, and gradually increased the amount of partially labeled data. We studied adding segmentation and type separately. After the model is trained in this routine, it was tested on predicting the full labels jointly on the test set. 

\paragraph{Domain Transfer with Partially Labeled Data}
In our final analysis we considered a novel domain-adaptation settings, where we have a small amount of  fully labeled in-domain data from aspect sentiment and more out-of-domain data from target sentiment. However unlike the traditional domain-adaptation settings, the out-of-domain data is labeled for a different task, and only shares one module with the original task.

\begin{figure}[t]
	\centering
	\includegraphics[width=0.31\textwidth]{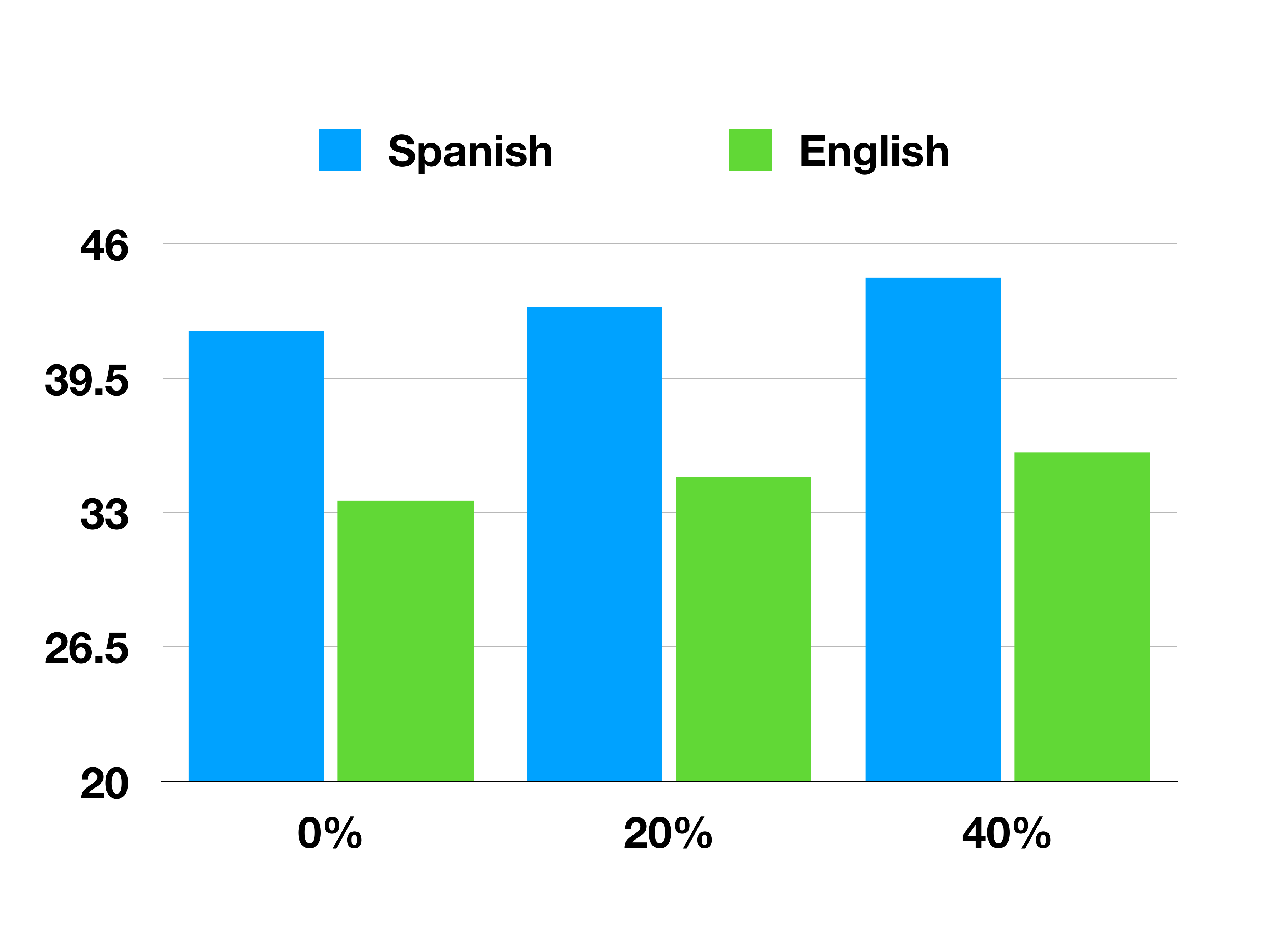}
	\caption{Domain Transfer experiments results with fixed $20\%$ in-domain data from aspect sentiment and varying amounts of out-of-domain data from target sentiment, shown on the x-axis.}
	\label{domaintransfer}
\end{figure}

In our experiments we fixed $20\%$ of the fully labeled data for the \textit{aspect} sentiment task, and gradually added  out-of-domain data, consisting of partial sentiment labels from the \textit{target} sentiment task. Our model successfully utilized the out-of-domain data and improved  performance on the in-domain task. The results are shown on Fig~\ref{domaintransfer}.

%% file: concl.tex
We present and study several modular neural architectures designed for a novel learning scenario: learning from partial labels. We experiment with several sentiment analysis tasks.
Our models, inspired by cognitive neuroscience findings~\cite{Jacobs1991,Eysenck2005} and multi-task learning, suggest a functional decomposition of the original task into two simpler sub-tasks. We evaluate different methods for sharing information and integrating the modules into the final decision, such that a better model can be learned, while converging faster\footnote{Convergence results are provided in the Appendix}.  
As our experiments show, modular learning can be used with weak supervision, using examples annotated with partial labels only. 

The modular approach also provides interesting directions for future research, focusing on alleviating the supervision bottleneck by using large amount of partially labeled data that are cheaper and easy to obtain, together with only a handful amount of annotated data, a scenario especially suitable for low-resource languages. 

%% file: appendix.tex
\section{Examples of Task Decomposition}
\begin{figure}[!hb]
\centering
\includegraphics[width=0.48\textwidth]{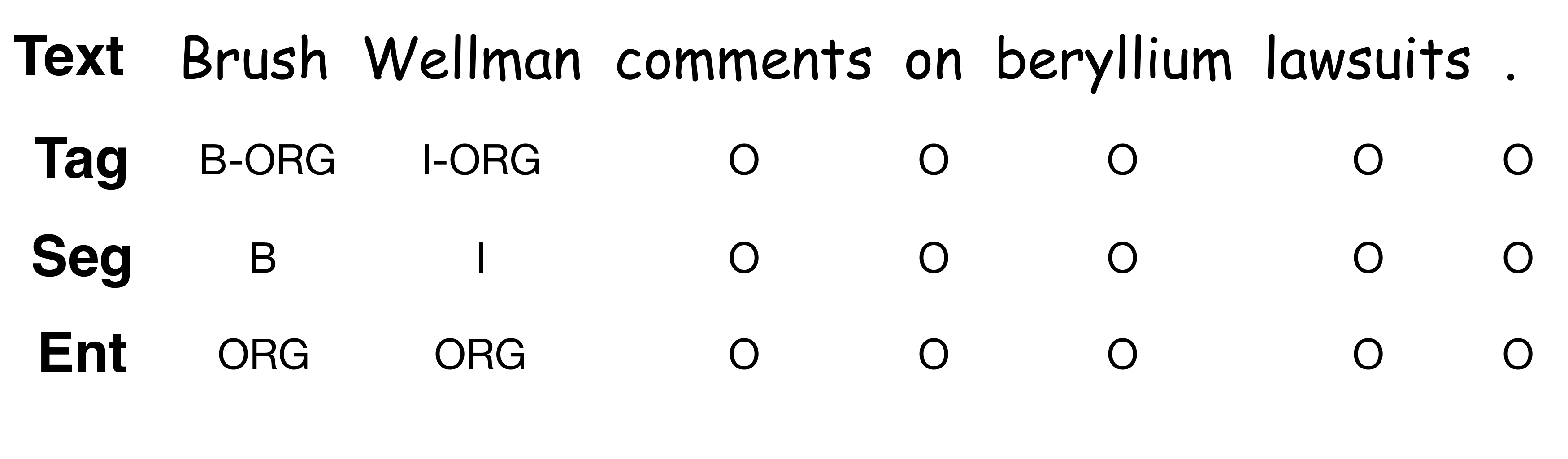}
	\caption{An example of NER decomposition.}
	\label{ner_eg}
\end{figure}

In Figure \ref{ner_eg}, we show an example of task decomposition for standard NER. 

\begin{figure}[!htb]
\centering
\includegraphics[width=0.48\textwidth]{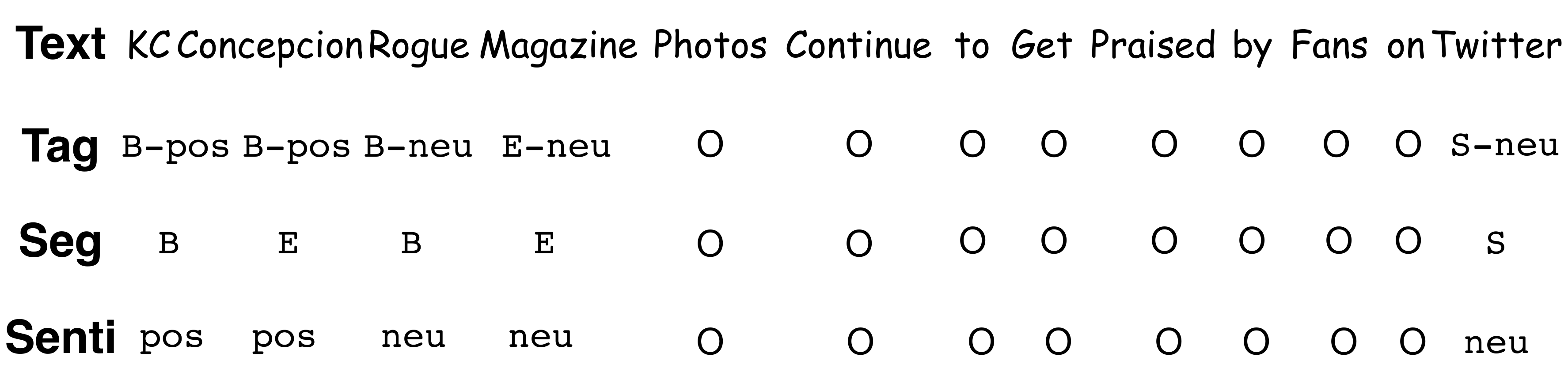}
	\caption{An extra example of target sentiment decomposition.}
	\label{ts_eg}
\end{figure}

In Figure \ref{ts_eg}, we show another example of task decomposition for target sentiment, in addition to the one in the main text.

\section{Full Experimental Results on Target Sentiment}
The complete results of our experiments on the target sentiment task are summarized in Tab.~\ref{target-senti-whole}. Our LSTM-CRF-TI(g) model outperforms all the other competing models in Precision, Recall and the F1 score.

\begin{table*}[!htb]
\small
\centering
\begin{tabular}{ |c|c||ccc|ccc| }
\hline
\multirow{2}{*}{System} & \multirow{2}{*}{Architecture} & \multicolumn{3}{c|}{English} & \multicolumn{3}{c|}{Spanish} \\ 
\cline{3-8} & & Pre & Rec & F1 & Pre & Rec & F1 \\ 
\hline
\multirow{4}{*}{Zhang, Zhang and Vo (2015)}
 & Pipeline & 43.71 & 37.12 & 40.06 & 45.99 & 40.57 & 43.04\\
 & Joint & 44.62 & 35.84 & 39.67 & 46.67 & 39.99 & 43.02\\
 & Collapsed & 46.32 & 32.84 & 38.36 & 47.69 & 34.53 & 40.00\\
 \hline
\multirow{4}{*}{Li and Lu (2017)} 
 & SS & 44.57 & 36.48 & 40.11 & 46.06 & 39.89 & 42.75\\
 & +embeddings & 47.30 & 40.36 & 43.55 & 47.14 & 41.48 & 44.13\\
 & +POS tags & 45.96 & 39.04 & 42.21 & 45.92 & 40.25 & 42.89\\
 & +semiMarkov & 44.49 & 37.93 & 40.94 & 44.12 & 40.34 & 42.14\\ 
 \hline
 \hline
 \multirow{1}{*}{Base Line} 
 & LSTM-CRF & 53.29 & 46.90 & 49.89 & 51.17 & 46.71 & 48.84 \\
 \hline
\multirow{3}{*}{\textit{This work}}
 & LSTM-CRF-T & 54.21 & 48.77 & 51.34 & 51.77 & 47.37 & 49.47\\
 & LSTM-CRF-Ti & 54.58 & 49.01 & 51.64 & 52.14 & 47.56 & 49.74\\
 & LSTM-CRF-Ti(g) & \textbf{55.31} & \textbf{49.36} & \textbf{52.15} & \textbf{52.82} & \textbf{48.41} & \textbf{50.50}\\
\hline
\end{tabular}
\caption{Performance on the target sentiment task}
\label{target-senti-whole}
\end{table*}

\section{Experiments on Named Entity Recognition}

\paragraph{NER datasets}
We evaluated our models on three NER datasets, the English, Dutch and Spanish parts of the 2002 and 2003 CoNLL shared tasks \cite{Sang2002,Sang2003}. We used the original division of training, validation and test sets. 
The task is defined over four different entity types: \textit{PERSON}, \textit{LOCATION}, \textit{ORGANIZATION}, \textit{MISC}. We used the \textsc{BIOES} tagging scheme during the training, and convert them back to original tagging scheme in testing as previous studies show that using this tagging scheme instead of \textsc{BIO2} can help improve performance~\cite{Ratinov2009,Lample2016,Ma2016,Liu2018}. As a result, the segmentation module had $5$ output labels, and the entity module had $4$. The final decision task, consisted of the Cartesian product of the segmentation set (BIES) and the entity set, plus the ``O'' tag, resulting in  $17$ labels.

\paragraph{Results on NER}
We compared our models with the state-of-the-art systems on English\footnote{\citeauthor{Liu2018}'s results are different since their implementation did not convert the predicted \textsc{BIOES} tags back to \textsc{BIO2} during evaluation. For fair comparison, we only report the results of the standard evaluation.}, Dutch and Spanish. For Dutch and Spanish, we used cross-lingual embedding as a way to exploit lexical information. The results are shown in Tab. \ref{NER-main} and Tab. \ref{compa_D_S}\footnote{We thank reviewers for pointing out a paper \cite{Agerri2016} obtains the new state-of-the-art result on Dutch with comparable results on Spanish.}. Our best-performing model outperform all the competing systems.
 

\begin{table}[!ht]
\small
\centering
\begin{tabular}{|l|c|}
\hline 
Model & English\\
\hline
LSTM-CRF~\cite{Lample2016} & 90.94\\ 
\hline
LSTM-CNN-CRF~\cite{Ma2016} & 91.21\\ 
\hline
LM-LSTM-CRF~\cite{Liu2018} & 91.06\\ 
\hline
\hline
LSTM-CRF-T & 90.8\\ 
\hline
LSTM-CRF-TI & 91.16\\ 
\hline
LSTM-CRF-TI(g) & \textbf{91.68}\\ 
\hline
\end{tabular}
\caption{Comparing our models with several state-of-the-art systems on the CoNLL 2003 English NER dataset.}
\label{NER-main}
\end{table}

\begin{table}[!ht]
\small
\centering
\begin{tabular}{|l|c|c|}
\hline 
Model & Dutch & Spanish\\
\hline
~\citeauthor{Carreras2002} \shortcite{Carreras2002} & 77.05 & 81.39\\
\hline
~\citeauthor{Nothman2013} \shortcite{Nothman2013}& 78.60 & N/A\\
\hline
~\citeauthor{Santos15} \shortcite{Santos15} & N/A & 82.21\\
\hline
~\citeauthor{Gillick2015} \shortcite{Gillick2015} & 82.84 & 82.95\\
\hline
~\citeauthor{Lample2016} \shortcite{Lample2016} & 81.74 & 85.75\\
\hline
\hline
LSTM-CRF-T & 83.91 & 84.89\\
\hline
LSTM-CRF-TI & 84.12 & 85.28\\
\hline
LSTM-CRF-TI(g) & \textbf{84.51} & \textbf{85.92}\\
\hline
\end{tabular}
\caption{Comparing our models with recent results on the 2002 CoNLL Dutch and Spanish NER datasets.}
\label{compa_D_S}
\end{table}

\section{Additional Experiments on Knowledge Integration}

\begin{figure}[!htb]
\begin{subfigure}{0.22\textwidth}
\centering
\includegraphics[scale=0.115]{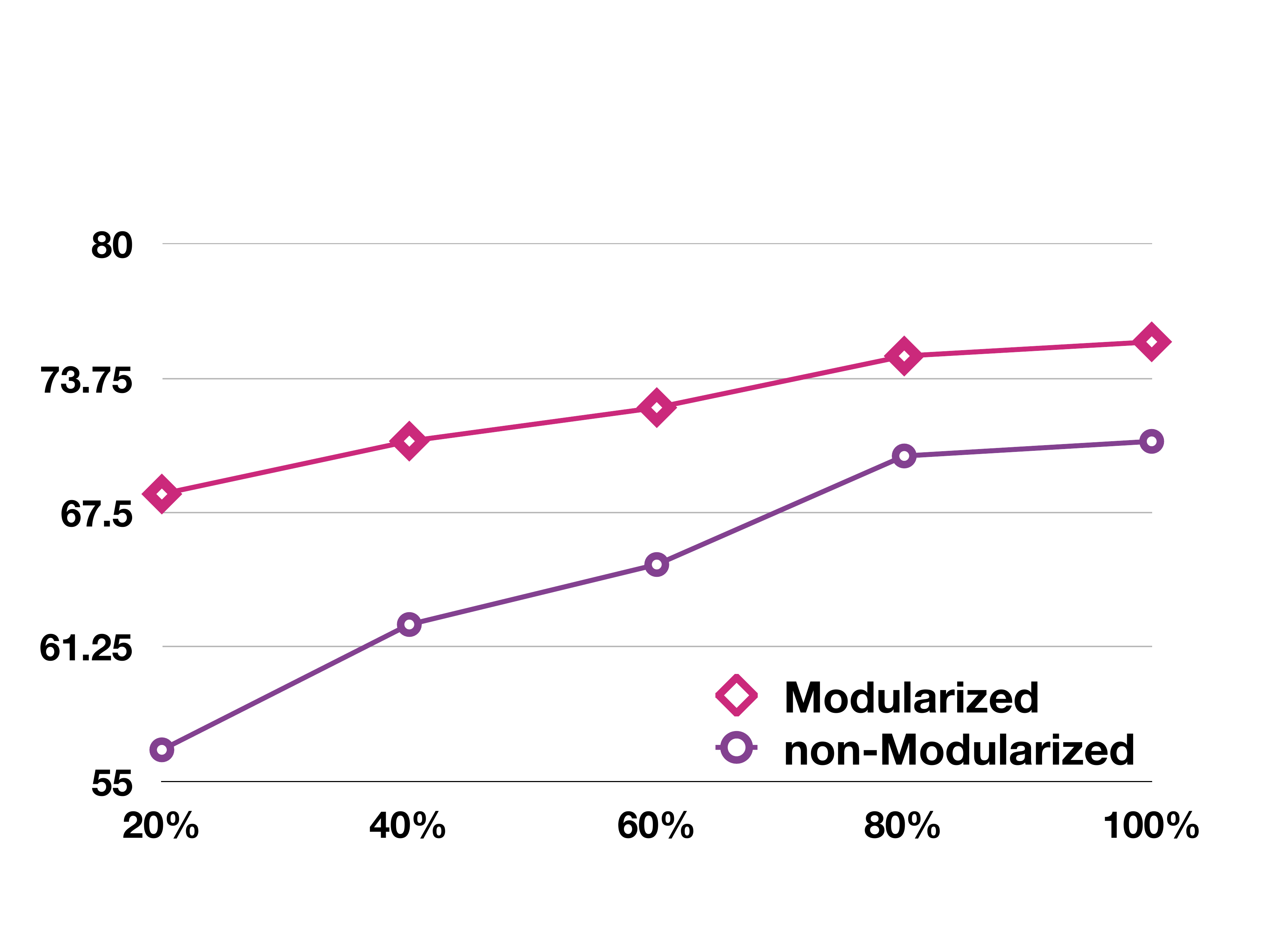}
\caption{Dutch NER}
\end{subfigure}
\quad
\begin{subfigure}{0.22\textwidth}
\centering
\includegraphics[scale=0.115]{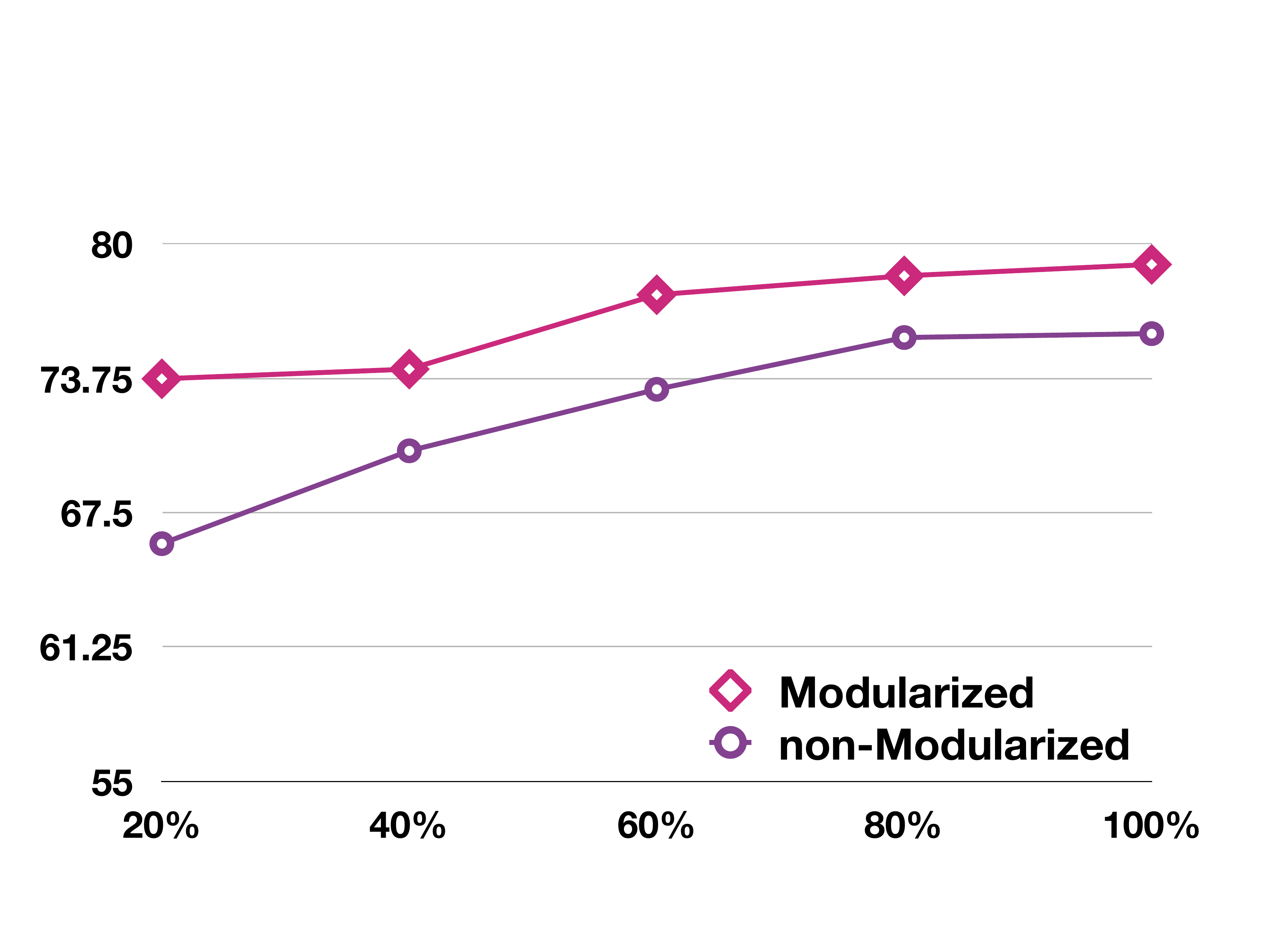}
\caption{Spanish NER}
\end{subfigure}
\caption{Experimental results on modular knowledge integration on the Dutch and Spanish NER datasets.}
\label{modular_ner}
\end{figure}

\begin{figure}[!htb]
	\centering
	\includegraphics[width=0.4\textwidth]{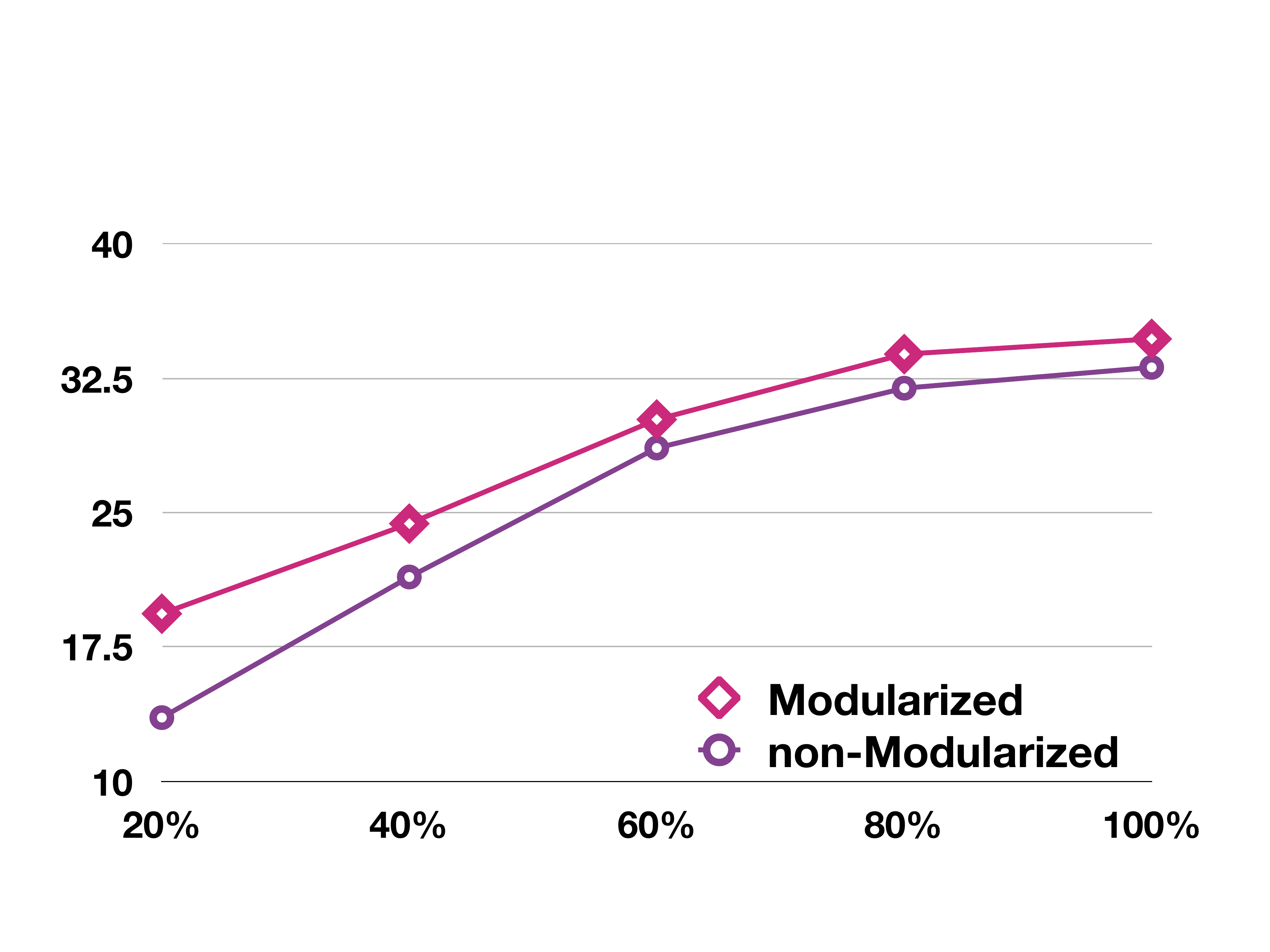}
	\caption{Experimental results on modular knowledge integration on the Subjective  Polarity  Disambiguation  Datasets.}
	\label{modular_sub}
\end{figure}

We conducted additional experiments on knowledge integration in the same setting as in the main text to investigate the properties of the modules. Figure~\ref{modular_ner} shows the results for Dutch and Spanish NER datasets, while Figure~\ref{modular_sub} shows the results for the Subjective  Polarity  Disambiguation  Datasets using the in-domain data.


\section{Convergence Analysis}
The proposed twofold modular infusion model (with guided gating as an option) breaks the complex learning problem into several sub-problems and then integrate them using joint training. The process defined by this formulation has more parameters and requires learning multiple objectives jointly. Our convergence analysis intends to evaluate whether the added complexity leads to a harder learning problem (i.e., slower to converge) or whether the tasks constrain each other and as a result can be efficiently learned.

\begin{figure}[!htb]
	\centering
	\includegraphics[width=0.4\textwidth]{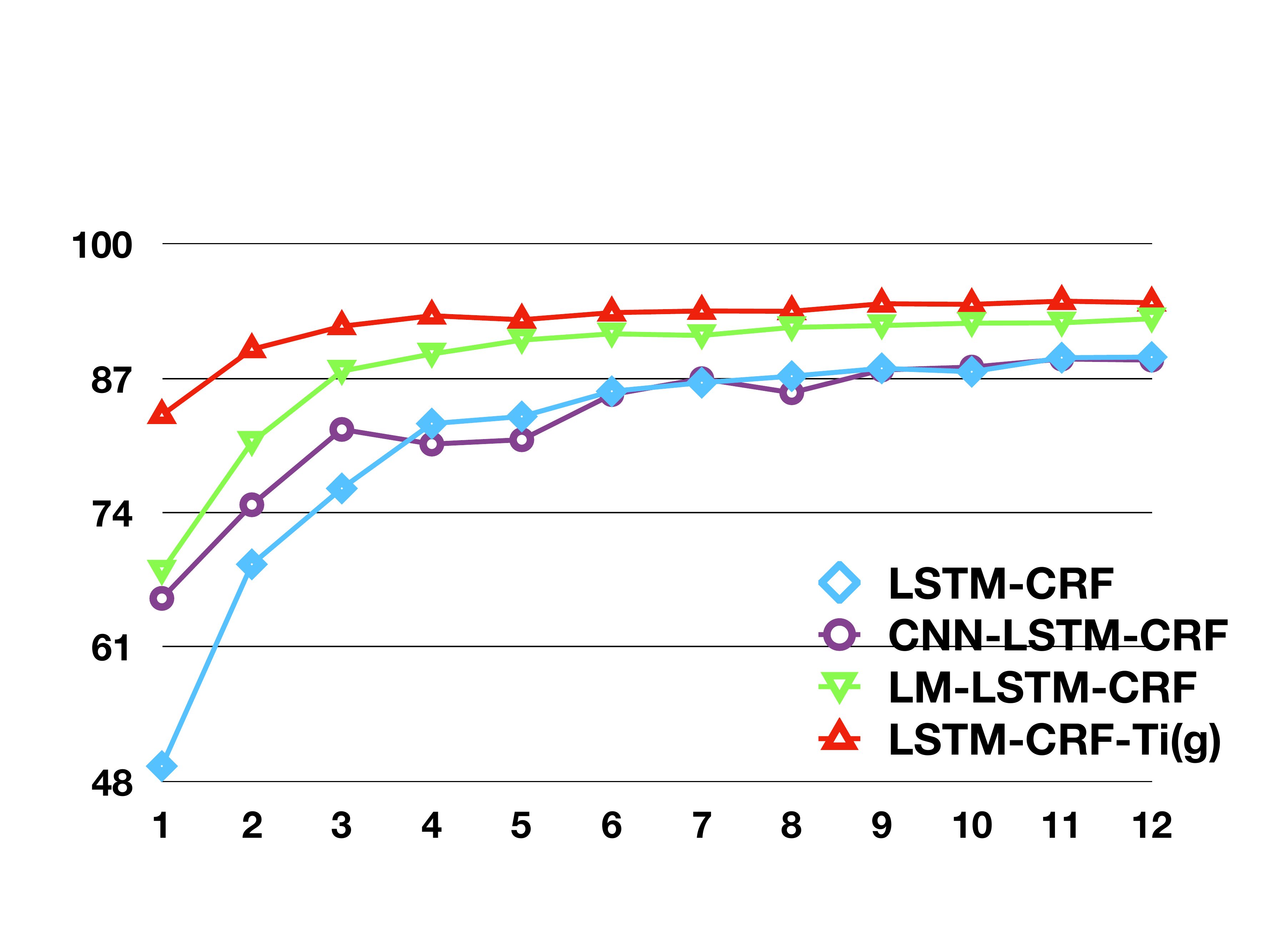}
	\caption{Comparing convergence over the development set on the English NER dataset. The x-axis is number of epochs and the y-axis is the F1-score.}
	\label{convergence}
\end{figure}

\begin{figure}[!htb]
	\centering
	\includegraphics[width=0.4\textwidth]{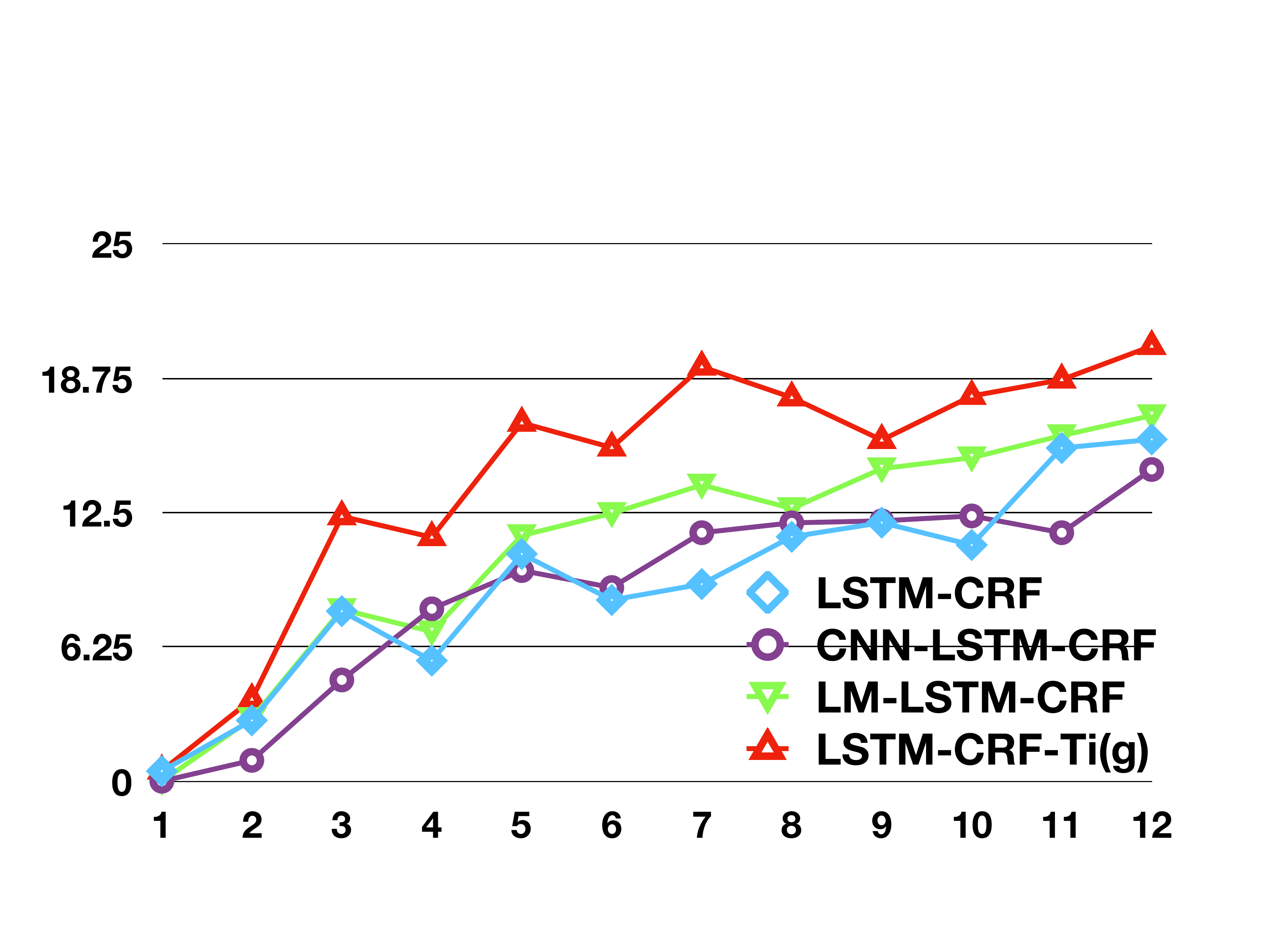}
	\caption{Comparing convergence over the development set on the subjective  polarity  disambiguation datasets. The x-axis is number of epochs and the y-axis is the F1-score.}
	\label{convergence-sub}
\end{figure}

We compare between our LSTM-CRF-TI(g) model and recent published top models on the English NER dataset in Figure \ref{convergence} and on the subjective  polarity  disambiguation datasets in Figure \ref{convergence-sub}. The curve compares convergence speed in terms of learning epochs. Our LSTM-CRF-TI(g) model has a much faster convergence rate compared to the other models. 




